\documentclass[journal, twocolumn, a4paper]{IEEEtran}

\usepackage{enumitem}
\usepackage{amsfonts}
\usepackage{pifont}
\usepackage{color,soul}
\usepackage{amssymb}

\usepackage{tikz}
\usepackage{pgfplots}
\usepackage{graphicx}
\usepackage{caption}
\usepackage{subcaption}

\usepackage{threeparttable}
\usepackage[hyphens]{url}

\usepackage{cite}
\usepackage[cmex10]{amsmath}
\usepackage{algorithmic}
\usepackage{algorithm}
\usepackage{array}
\usepackage{xifthen}

\newcommand{\vbIfArgNEmp}[2]{\ifthenelse{\isempty{#1}}{}{#2}}

\newcommand{\StateVec}[2]{
  \ensuremath{\mathbf{x}\vbIfArgNEmp{#1}{^{#1}}\vbIfArgNEmp{#2}{_{#2}}}
}

\newcommand{\MeasVec}[2]{
  \ensuremath{\mathbf{z}\vbIfArgNEmp{#1}{^{#1}}\vbIfArgNEmp{#2}{_{#2}}}
}

\newcommand{\MeasVecTrj}[3]{
  \ensuremath{\mathbf{z}\vbIfArgNEmp{#1}{^{#1}}_{\vbIfArgNEmp{#2}{{#2}}:\vbIfArgNEmp{#3}{{#3}}}}
}

\newlength\figureheight
\newlength\figurewidth

\begin{document}

\title{Incremental Nonlinear System Identification  and
Adaptive Particle Filtering Using Gaussian Process}

\author{Vahid~Bastani,~\IEEEmembership{Student,~IEEE,}
  Lucio~Marcenaro,~\IEEEmembership{Member,~IEEE,}
  and~Carlo~S.~Regazzoni,~\IEEEmembership{Senior Member,~IEEE}
  \thanks{V. Bastani. L. Marcenaro and C. S. Regazzoni are with the Department of Electrical, Electronics and Telecommunication Engineering and Naval Architecture (DITEN), University of Genova, 16145, Via All'Opera Pia 11A, Geona, Italy ( email: vahid.bastani@ginevra.dibe.unige.it, lucio.marcenaro@unige.it, carlo.regazzoni@unige.it)}
}

\markboth{}%
{Shell \MakeLowercase{\textit{et al.}}: Bare Demo of IEEEtran.cls for Journals}

\maketitle

\begin{abstract}
  An incremental/online state dynamic learning method is proposed for
identification of the nonlinear Gaussian state space models. The method embeds
the stochastic variational sparse Gaussian process as the probabilistc state
dynamic model inside a particle filter framework. Model updating is done at
measurement sample rate using stochastic gradient descent based optimisation
implemented in the state estimation filtering loop. The performance of the
proposed method is compared with state-of-the-art Gussian process based batch
learning methods.  Finally, it is shown that the state estimation performance
significantly improves due to the online learning of state dynamics.

\end{abstract}


\IEEEpeerreviewmaketitle

\section{Introduction}
\label{sec:intro}
\IEEEPARstart{B}{a}yesian filtering (BF) is the most widespread technique for
state estimation in science and engineering. It has been used in many diverse
fields including but not limited to signal processing, computer vision, control,
robotic and economy. BF requires that the dynamics of the state of the system be
known up to some tolerable uncertainty. The fundamental difficulty of BF is to
find a correct stochastic process model of the dynamics of the system.
Failing to specify a correct and justifiable model will
severely impacts the performance of BF and puts it in the risk of
undetectable arbitrarily large error.

Linear dynamic model is the commonly used classical model. In this case, Kalman
Filter provides efficient and fast solution for BF. However, in the majority of
real world applications, the dynamics are nonlinear. Moreover, in the linear
models the parameters has to carefully be chosen \cite{heffes1966} as well.
Particle filtering (PF) is the most flexible form of the BF based on sequential
Monte-Carlo that can be applied on nonlinear non-Gaussian dynamic models. Having
a correct model in the PF is even more crucial as the PF highly relies on the
state dynamic model for sampling process. Filtering under dynamic model
uncertainity has been studied in \cite{ardeshiri2015, ghahramani1996} for linear
dynamic systems, \cite{ozkan2015, nemeth2014, erol2013, chopin2013} for
parametric state space models.

In this paper, an incremental/online nonparametric method is proposed for
learning nonlinear dynamics in state space model.
The Gaussian Process (GP) regression is used here for learning the nonlinear
function that models the state dynamic. Incremental model updating is achieved
using the stochastic variational inference of GP. The model updating is
integrated inside a PF loop. The proposed method is particularly
useful when the measurement data is received in sequence and there is no
training data available for learning. Furthermore, when a
large number of data is available, it is only practical to process data in
sequences or small batches due to the computational resource constraints. One
immediate application of learned model is for BF state estimation. This is shown
in this paper, where the performance of the PF used in the proposed framework
increases gradually since it uses the incrementally learned model for sampling
process. However, the learned model can also be used for classification and
abnormality detection purposes \cite{bastani2016, bastani2015}. Simulating
similar data is another application of the learned model which can be used for
state prediction as well.

The paper is organized as follows: Section \ref{sec:ssm} defines the nonlinear
state space model. In Section \ref{sec:incremental} the proposed incremental
model identification algorithm is presented. In Section \ref{sec:eval} the
performance of the proposed technique is analysed and compared with
the state-of-the-art. Finally, Section \ref{sec:concl} concludes the paper.

\section{Nonlinear State Space Model}
\label{sec:ssm}
The state space model (SSM) of a dynamic system is defined using three random
processes:
\begin{equation}
  \begin{aligned}
    \StateVec{}{0} &\sim p_0(\StateVec{}{0})
    \\
    \StateVec{}{t}|\StateVec{}{t-1} &\sim p_f(\StateVec{}{t}|\StateVec{}{t-1})
    \\
    \MeasVec{}{t}|\StateVec{}{t} &\sim p_g(\MeasVec{}{t}|\StateVec{}{t}),
  \end {aligned}
\label{equ:ssmp}
\end {equation}
where $\StateVec{}{t}$ and $\MeasVec{}{t}$ are the state and measurement
vectors at time $t$, $p_0$ is the initial state probability distribution
function (PDF), $p_f$ is a conditional probability density function (CPDF)
representing the dynamics of the state and $p_g$ is a CPDF representing the
measurement process. In a Gaussian nonlinear system the above CPDFs are
constructed by:
\begin{equation}
  \begin{aligned}
    \StateVec{}{t} =& f(\StateVec{}{t-1}) + \boldsymbol\omega_t
    \\
    \MeasVec{}{t} =& g(\StateVec{}{t}) + \boldsymbol\nu_t,
\end {aligned}
\label{equ:ssme}
\end {equation}
where $f$ and $g$ are nonlinear functions and $\boldsymbol\omega_t$ and
$\boldsymbol\nu_t$ are zero-mean white Gaussian noises. Conventional state
estimation problem considers estimating the posterior of state sequence
$\{\StateVec{}{0}, \cdots, \StateVec{}{t}\}$ given the measurement sequence
$\{\MeasVec{}{1}, \cdots, \MeasVec{}{t}\}$ while all other parameters of the
system are known. In BF this is achieved by recursively calculating the filtered
state posterior:
\begin{equation}
  \begin{aligned}
    &p(\StateVec{}{t}|\StateVec{}{t-1}, \cdots, \StateVec{}{0}, \MeasVec{}{t},
  \cdots, \MeasVec{}{1}) \propto \\
  &p_f(\StateVec{}{t}|\StateVec{}{t-1})
   p_g(\MeasVec{}{t}|\StateVec{}{t})
  p(\StateVec{}{t-1}|\StateVec{}{t-2}, \cdots, \StateVec{}{0}, \MeasVec{}{t-1},
  \cdots, \MeasVec{}{1}).
  \end{aligned}
\end{equation}
 This paper deals with the state estimation problem when the
dynamic model $f$ is unknown. The goal is to estimate jointly the state sequence
and $f$ from measurement sequence. However, the presented technique can be used
for estimating $g$ while $f$ is known. Note that when both $f$ and $g$ are
unknown the problem is highly ill-posed and can only be attempted with sensible
constraints.

\section {Incremental Model Identification}
\label{sec:incremental}
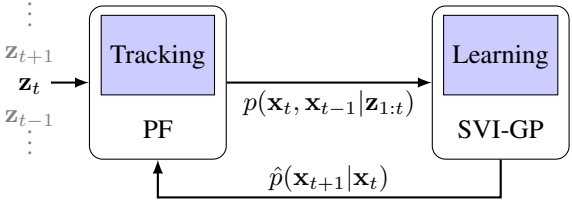
\begin{figure}
  \usetikzlibrary{shapes,arrows}

\usetikzlibrary{fit}					
\usetikzlibrary{backgrounds}	
\usetikzlibrary{positioning,calc}

\centering

\tikzstyle{area}=[rectangle, rounded corners=5pt, draw=black, minimum size=1.8cm ]

\tikzstyle{block} = [draw, fill=blue!20, rectangle,   minimum height=3em]

\begin{tikzpicture}[>=latex,text height=1.5ex,text depth=0.25ex]

\node [block](blktrack){Tracking};
\node [below of = blktrack](b1label){PF};

\node [block, right= 3cm of blktrack] (blklear){Learning};
\node [below of = blklear](b2label){SVI-GP};

\begin{pgfonlayer}{background}
  \node (bk1)[area, fit={(blktrack)(b1label)}] {};
	\node (bk2)[area, fit={(blklear)(b2label)}] {};
\end{pgfonlayer}

\draw [thick, ->]
  (bk2.south) -- 
  ++(0cm, -0.5cm) --
  node[above, pos=0.5]{$\hat{p}(\StateVec{}{t+1}|\StateVec{}{t})$} 
  ($(bk1.south)+(0cm, -0.5cm)$) -|
  (bk1.south);

\node[left = .5 cm of bk1](inp) {$\MeasVec{}{t} $};
\node[above = -1mm of inp, text=gray](inpp) {$\MeasVec{}{t+1} $};
\node[below = -1mm of inp, text=gray](inpm) {$\MeasVec{}{t-1} $};
\node[below = -1mm of inpm, text=gray] {$\vdots$};
\node[above = -1mm of inpp, text=gray] {$\vdots$};

\draw [thick,->] (bk1) --
  node[below, pos=0.5]{$p(\StateVec{}{t},\StateVec{}{t-1}|\MeasVecTrj{}{1}{t})$}
  (bk2);

\draw [thick,->] (inp) -- (bk1);


\end{tikzpicture}
  \caption{Simplified diagram of incremental dynamic model identification.}
  \label{fig:loop}
\end{figure}

Fig. \ref{fig:loop} shows a simplified diagram of the proposed incremental
identification problem. At the instance $t$ the measurement $\MeasVec{}{t}$ is
received. The block \textit{Tracker} uses the measurement and the current
estimate of the state dynamic model to produce a joint posterior distribution of the
current state $\StateVec{}{t}$ and the previous state $\StateVec{}{t-1}$. The
posterior is then fed to \textit{Learning} block that uses it for updating the
estimate of the state dynamic model.

Due to the nonlinear settings of the problem, the conventional Sequential
Importance Resampling (SIR) PF \cite{arulampalam2002} is used here as \textit{Tracker}.
In this case then posterior
$p(\StateVec{}{t},\StateVec{}{t-1}|\MeasVecTrj{}{1}{t})$ is given as a set of
$N$ weighted particles
$\{\StateVec{(i)}{t},\StateVec{(i)}{t-1}, \omega^{(i)}\}_{i=1}^N$ with
$\omega^{(i)}$ denotes the weight of $i^{\text{th}}$ particle. Note that,
for the PF algorithm it is only necessary to keep the particle of the
current state. However as $\StateVec{}{t}$ and $\StateVec{}{t-1}$
are domain and codomain of the
function $f$, it is necessary to jointly estimate both which are then used in
\textit{Learning} block. This is achieved by simply keeping particles of previous $t-1$
iteration in the memory. That makes $\StateVec{(i)}{t}$ the filtered state
particle and the $\StateVec{(i)}{t-1}$ the one-step-lag smoothed state particle.

The \textit{Learning} process incrementally updates the probability model $p$ at
each step and provides the updated model to the PF. Stochastic-Variational
Sparse Gaussian Process (SVSGP) \cite{hensman13} is used here as
\textit{Learning} mechanism.  Gaussian Process \cite{rasmussen2006} model is a
well established Bayesian nonparametric function regression technique. Its
ability for capturing and propagating uncertainties from the training samples to
the posterior regression model makes it perfectly fit in the Bayesian
framework. 
 
\subsection {Stochastic Variational Gaussian Process}
\label{subsec:svsgp}

A Gaussian Process (GP) defines a probability distribution over functions $f:
\mathcal{X} \rightarrow \mathbb{R}$ such that the marginal distribution of
vectorized function values $\Gamma = [f(\StateVec{}{1}), \cdots,
f(\StateVec{}{N})]^T$ over any finite subset $\{\StateVec{}{1}, \cdots,
\StateVec{}{N}\} \subset \mathcal{X}$ be a multivariate Gaussian
\cite{rasmussen2006}. A GP, denoted $f(\StateVec{}{}) \sim
GP(\bar{f}(\StateVec{}{}), k(\StateVec{}{},\StateVec{}{}'))$, is characterized
by a mean function $\bar{f}(\StateVec{}{})$ and a covariance function
$k(\StateVec{}{},\StateVec{}{}')$ that encodes covariance of two values,
$f(\StateVec{}{})$ and $f(\StateVec{}{}')$.

The GP has widely been applied in Bayesian nonlinear, nonparametric regression
problems. Consider training data set $\mathcal{D} = \{X,\mathbf{y}\}$ consists
of noisy function values  $\mathbf{y} = [y_1, \cdots,y_n]^T$ at the set of
points $X = \{\StateVec{}{1}, \cdots, \StateVec{}{N}\}$, where $y_i =
f(\StateVec{}{i})+e$ and $e$ is a white Gaussian noise.  The GP regression
considers estimating the values of function $\mathbf{f}^*= [f(\StateVec{*}{1}),
\cdots, f(\StateVec{*}{M})]^T$ at a set of new points $X^* = \{\StateVec{*}{1},
\cdots, \StateVec{*}{M}\}$ where $f(\StateVec{}{})$ has a GP prior. The
posterior $p(\mathbf{f}^*|\mathbf{y})$ is a Gaussian with mean vector:
\begin{equation}
  \mathbf{K}_{MN}[\mathbf{K}_{NN}+\sigma\mathbf{I}]^{-1}\mathbf{y}.
\end{equation}
and covariance matrix
\begin{equation}
  \mathbf{K}_{MM}-\mathbf{K}_{MN}[\mathbf{K}_{NN}+\sigma\mathbf{I}]^{-1}\mathbf{K}_{NM}.
\end{equation}
where $\mathbf{K}_{MM}$, $\mathbf{K}_{MN}$, $\mathbf{K}_{NN}$ and
$\mathbf{K}_{NM}$ are covariance matrices whose elements are
$k(\StateVec{*}{i},\StateVec{*}{j})$, 
$k(\StateVec{*}{i},\StateVec{}{j})$,
$k(\StateVec{}{i},\StateVec{}{j})$ and
$k(\StateVec{}{i},\StateVec{*}{j})$
respectively.  The covariance function and the noise variance control the poster
GP. These hyper-parameters are optimized by maximizing the training data
marginal log likelihood.
\begin{equation}
  \log  p(\mathbf{y}) = \log \mathcal{N}(\bar{\mathbf{f}},
  \sigma^2\mathbf{I}+\mathbf{K}_{NN}).
\label{eq:gplik}
\end{equation}

The inference in the standard GP has $\mathcal{O}(N^2)$ memory demand and
$\mathcal{O}(N^3)$ time complexity. Sparse variational GP \cite{titsias09}
reduces complexity by approximating the data set using a variational
distribution $q(\mathbf{u}) = \mathcal{N}(\mathbf{m}, \mathbf{S})$ representing
the function values over set of inducing points $Z = \{\mathbf{z}_1, \cdots,
\mathbf{z}_{L}\}$ that maximize the variational lower-bound of
(\ref{eq:gplik}):
\begin{equation}
  p(\mathbf{y}|Z) >   \log \mathcal{N}(\bar{\mathbf{f}},
  \sigma^2\mathbf{I}+\mathbf{K}_{N L}\mathbf{K}_{L L}^{-1}\mathbf{K}_{L N})
  \triangleq \mathcal{L}
\label{eq:sgplower}
\end{equation}
This way the memory and complexity of the inference task will reduce to
$\mathcal{O}(NM)$ and $\mathcal{O}(NM^2)$. This can still be prohibitive for
\textit{Big Data} problem where $N$ is large. Stochastic Variational Sparse
Gaussian Process (SVSGP) \cite{hensman13} proposes another lower bound: 
\begin{equation}
\begin{split}
  &\mathcal{L} \geq \mathcal{L}' \triangleq \\
  &\sum\limits_{i=1}^N 
  \Big\{\log \mathcal{N}(y_i|\bar{f}_i+\mathbf{k}_i^T\mathbf{K}_{LL}^{-1}\mathbf{m},
  \sigma)-\frac{\tilde{k}_{i,i}}{2\sigma}-\frac{\text{tr}(\mathbf{S}\boldsymbol\Lambda_i)}{2}\Big\}\\
  &- \mathcal{D}_{KL}(q(\mathbf{u})||p(\mathbf{u})).
\end{split}
\label{eq:svsgplower}
\end{equation}
where $\mathbf{k}_i$ is the $i^{\text{th}}$ column of $\mathbf{K}_{L N}$,
$\boldsymbol\Lambda_i =
\sigma^{-1}\mathbf{K}_{LL}^{-1}\mathbf{k}_i\mathbf{k}_i^T\mathbf{K}_{LL}^{-1}$
and $\tilde{k}_{i,i}$ is the $i^{\text{th}}$ diagonal of
$\mathbf{K}_{NN}-\mathbf{K}_{NL}\mathbf{K}_{LL}^{-1}\mathbf{K}_{LN}$. The
difference between $\mathcal{L}$ and $\mathcal{L}'$ is that in the latter the
variational distribution parameters are explicit while in the former they are
analytically optimized out. However, $\mathcal{L}'$ is written as $N$ terms
corresponding to each training data pair. This is the necessary condition for
the objective function of stochastic gradient descent (SGD) optimization. The
SGD uses approximate gradient from mini-batch in each iteration of gradient
descent instead of full gradient calculated on the whole dataset.

The training of SVSGP is done by taking steps in the direction of approximate
gradient in each iteration. Since the approximate gradient is calculated on a
subset of training data it is possible to use this in online learning. In online
learning the training data is received one by one or in small batches from a
supposedly infinite length process.

\subsection {Domain Variable Uncertainty in GP}
\label{subsec:uncertainty}
The standard GP regression assumes training inputs domain are noiseless. This is
not the case here as the output of the PF is an estimated joint distribution
$\hat{p}(\StateVec{}{t}, \StateVec{}{t-1})$ of codomain-domain variables of the GP. 
domain variable uncertainty in GP has been addressed in \cite{mchutchon11} for
special case of Gaussian i.i.d noise. However, this is not applicable
in the problem of this paper as the joint distribution may take any form in
the nonlinear dynamics.

A trivial solution is to use particle pairs
$\{\StateVec{(i)}{t},\StateVec{(i)}{t-1}\}_{i=1}^N$ as data mini-batches for
SVSGP training.  However, as the SVGP values all the training data the same and
the weights are ignored, this solution is highly inefficient. Alternatively, one
may approximate the distribution $\hat{p}(\StateVec{}{t}, \StateVec{}{t-1})
=\sum \omega^{i}
\delta(\StateVec{}{t}-\StateVec{(i)}{t},\StateVec{}{t-1}-\StateVec{(i)}{t-1})$
with a uniformly weighted particle distribution $\hat{q}(\StateVec{}{t},
\StateVec{}{t-1}) =\frac{1}{N}\sum
\delta(\StateVec{}{t}-\tilde{\mathbf{x}}^{(i)}_t,\StateVec{}{t-1}-\tilde{\mathbf{x}}^{(i)}_{t-1})$
and use the equally weighted particles set
$\{\tilde{\mathbf{x}}^{(i)}_t,\tilde{\mathbf{x}}^{(i)}_{t-1}\}$ as
mini-batches for GP training.  $\hat{q}$ can be optimized by minimizing the KL
divergence:
\begin{equation}
  KL(\hat{p}||\hat{q}) = \sum \omega^{(i)} \log\frac{N\omega^{(i)}}{\eta_i}
  \label{eq:resampling}
\end{equation}
subject to $\sum \eta_i = N$ and $\eta_i \in \mathbb{N}$, where $\eta_i$ is the number of elements in
$\{(\tilde{\mathbf{x}}^{(j)}_t,\tilde{\mathbf{x}}^{(j)}_{t-1})\}_{j=1}^N$ 
that are equal to $(\StateVec{(i)}{t},\StateVec{(i)}{t-1})$. It is easy to
verify that the $\eta_i$ that solves (\ref{eq:resampling}) have to be
approximately proportional to $\omega^{(i)}$. In fact, solving for $\hat{q}$ is
exactly equivalent to resampling process in the PF for particle degeneracy
mitigation \cite{arulampalam2002}. 

Resampling replicates particles with larger weights and removes low weight
particles. Using resampled particles for GP training artificially incorporates
their weights since the contribution of each particle get multiplied
proportional to its weights due to the summation in GP objective function
(\ref{eq:svsgplower}).

\subsection {The Algorithm}
\label{subsec:algo}
Algorithm \ref{alg:alg} shows one iteration of the proposed method. 
$\hat{\sigma}_{t}$, $\hat{\theta}_{t}$, $\hat{\mathbf{m}}_{t}$ and $\hat{\mathbf{S}}_{t}$
denote estimated dynamic noise variance, parameter of GP kernel, mean of
$q$ and covariance of $q$ respectively after $t^{\text{th}}$ measurement. The
gradient descend step $GD(\cdots)$ is done by in the standard way. 

\begin{algorithm}[H]
  \caption{An iteration of incremental model identification}
  \label{alg:alg}
  Input: $\MeasVec{}{t}$, $\{\StateVec{(i)}{t-1},\StateVec{(i)}{t-2},
  \omega^{(i)}\}_{i=1}^N$, $\hat{\sigma}_{t-1}$, $\hat{\theta}_{t-1}$,
  $\hat{\mathbf{m}}_{t-1}$, $\hat{\mathbf{S}}_{t-1}$

Output: $\{\StateVec{(i)}{t},\StateVec{(i)}{t-1},
  \omega^{(i)}\}_{i=1}^N$, $\hat{\sigma}_{t}$, $\hat{\theta}_{t}$,
  $\hat{\mathbf{m}}_{t}$, $\hat{\mathbf{S}}_{t}$

\begin{enumerate}
  \item Optionally resample $\{\StateVec{(i)}{t-1},\StateVec{(i)}{t-2},
  \omega^{(i)}\}_{i=1}^N$ to avoid degeneracy.
  \item Sample $\StateVec{(i)}{t} \sim
    \hat{p}_f(\StateVec{}{t}|\StateVec{(i)}{t-1})$ for $i=1, \cdots, N$.
  \item Let $\omega^{(i)} = \omega^{(i)}p_g(\MeasVec{}{t}|\StateVec{(i)}{t})$
    for $i=1, \cdots, N$.
  \item Resample $\{\StateVec{(i)}{t},\StateVec{(i)}{t-1},
    \omega^{(i)}\}_{i=1}^N$ to
    $\{\tilde{\mathbf{x}}^{(j)}_t,\tilde{\mathbf{x}}^{(j)}_{t-1}\}_{j=1}^N$ to
    minimize (\ref{eq:resampling}).
  \item Calculate
    $\mathcal{L}'$ for $\{\tilde{\mathbf{x}}^{(j)}_t,\tilde{\mathbf{x}}^{(j)}_{t-1}\}_{j=1}^N$
    from (\ref{eq:svsgplower}).
  \item Calculate gradient $\nabla \mathcal{L}' = [\frac{\partial \mathcal{L}'}{\partial \hat{\sigma}_{t-1}},
    \frac{\partial \mathcal{L}'}{\hat{\theta}_{t-1}},
    \frac{\partial \mathcal{L}'}{\hat{\mathbf{m}}_{t-1}},
  \frac{\partial \mathcal{L}'}{\hat{\mathbf{S}}_{t-1}}]$
  for $\{\tilde{\mathbf{x}}^{(j)}_t,\tilde{\mathbf{x}}^{(j)}_{t-1}\}_{j=1}^N$
  \cite{hensman13}.
\item Calculate new parameters using gradient descend:
  $\hat{\sigma}_{t}, \hat{\theta}_{t},
  \hat{\mathbf{m}}_{t}, \hat{\mathbf{S}}_{t} \leftarrow GD(\mathcal{L}',
  \nabla \mathcal{L}', \hat{\sigma}_{t-1}, \hat{\theta}_{t-1},
  \hat{\mathbf{m}}_{t-1}, \hat{\mathbf{S}}_{t-1})$

\end{enumerate}

\end{algorithm}

\section {Evaluation}
\label{sec:eval}
\subsection{Comparison}
\label{ssec:comp}
The performance of the proposed method is compared with GP-SSM
\cite{frigola14} and GP-NARX  \cite{candela2003} which are both GP-based. Unlike
proposed method, these two methods are batch based that is working on full training data.
It should be noted that \cite{frigola14} also proposes a
stochastic variational inference and discusses possible online application, but it
is left without elaboration. The same evaluation setup in
\cite{frigola14} is used here for comparison. The algorithms applied on the samples of a
nonlinear dynamic model defined by $p(x_t|x_{t-1})=\mathcal{N}(f(x_{t-1}), 1)$
and $p(z_t|x_t) = \mathcal{N}(x_t, 1)$ where
\begin{equation}
  f(x) =
  \begin{cases}
    x+1 \quad x < 4, \\
    -4x+21 \quad x \geq 4.
  \end{cases}
\label{eq:testfunc}
\end{equation}

\begin{figure}
  \setlength\figureheight{3.3cm} 
  \setlength\figurewidth{7cm}
  \input{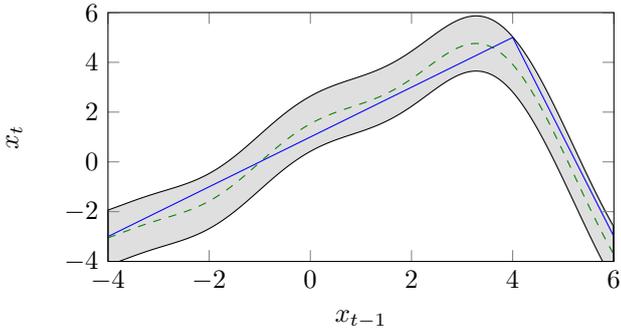}
  \caption{The test function and the output of the GP trained with proposed
  algorithm.}
  \label{fig:testfunc}
\end{figure}

Table \ref{tbl:comp} compares the performances of the proposed method with the
state-of-the-art. The methods are trained with a sequence of $500$ samples then
they are tested with another sequence of $10^4$ samples. The Mat\`{e}rn kernel
is used for all GP based algorithms. Fig. \ref{fig:testfunc} shows the test
function and the function learned by the proposed method. The performance
metrics are the Mean Squared Error (MSE) between the test samples and the
predictions and the Mean Log Likelihood (MLL) of the test samples given the
trained model $p(x_{t}^{\text{test}}|x_{t-1}^{\text{test}})$.  As the Table
\ref{tbl:comp} shows, despite the proposed method is incremental/online, its
performance is comparable to the state-of-the-art.  GP-SSM. The MSE is slightly
higher than the GP-SSM while the MLL is improved a little.

\begin{table}[h]
\caption{Learning performance comparison}
\begin{center}
\begin{threeparttable}
	\begin{tabular}{lll}
    Method & Test MSE & Test MLL \\
   	\hline
    Proposed (incremental)  &  $1.17$ & $-1.56$\\
    SSM-GP (batch) & $1.15$ & $-1.61$\\
    GP-NARX (batch) & $1.46$ & $-1.90$ \\
		\hline
	\end{tabular}
\end{threeparttable}
\end{center}
\label{tbl:comp}
\end{table}

\subsection{Performance}
\label{ssec:pef}
The incremental learning performance of proposed method is evaluated using
simulated nonlinear dynamic models given as

\begin{figure}[t]
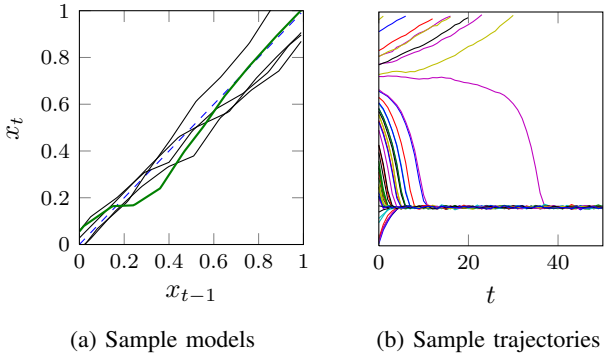

  \begin{subfigure}[t]{0.49\columnwidth}
    \centering
    \setlength\figureheight{3.1cm} 
    \setlength\figurewidth{3.1cm}
    \input{./fig/models.tex}
    \caption{Sample models}
    \label{fig:models}
  \end{subfigure}
  \begin{subfigure}[t]{0.45\columnwidth}
    \centering
    \setlength\figureheight{3.1cm} 
    \setlength\figurewidth{3.1cm}
    \input{./fig/trajs.tex}
    \caption{Sample trajectories}
    \label{fig:trajs}
  \end{subfigure}
  
  \caption{Some samples of simulated dynamic models (a) and trajectories (b)
  used for evaluation}
\end{figure}

$p(x_t|x_{t-1})=\mathcal{N}(f(x_{t-1}), 10^{-2})$ and $p(z_t|x_t) =
\mathcal{N}(x_t, 10^{-3})$ where
\begin{equation}
  f(x) = x+
  \begin{cases}
    \frac{b_{1}-b_{0}}{a_1-a_{0}}(x-a_{0})  & a_{0} \leq x < a_{1} \\
    \quad \vdots & \quad \vdots \\
    \frac{b_{n}-b_{n-1}}{a_n-a_{n-1}}(x-a_{n-1}) & a_{n-1} \leq x < a_{n} \\
  \end{cases}
\label{eq:rndfunc}
\end{equation}
with $(a_{i} - a_{i-1}) \sim \mathcal{U}(0.08, 0.15)$, $(b_{i} - b_{i-1}) \sim
\mathcal{N}(0, 10^{-3})$ and $n = 20$. Unlike (\ref{eq:testfunc}),
(\ref{eq:rndfunc}) generates smooth trajectories which are more realistic
as systems are usually constrained by energy. $50$ random
functions are generated from (\ref{eq:rndfunc}) by sampling $a_i$ and $b_i$.
Five samples of such function are shown in Fig. \ref{fig:models}. Using each
random function $50$ trajectories are simulated with $p(x_0) = \mathcal{U}(0,
1)$. The models are producing diverse trajectory shapes. 
Fig. \ref{fig:trajs} shows sample trajectories generated by
the highlighted function in Fig. \ref{fig:models}.

\begin{figure}[t]
    \setlength\figureheight{3.3cm} 
    \setlength\figurewidth{7cm}
  \input{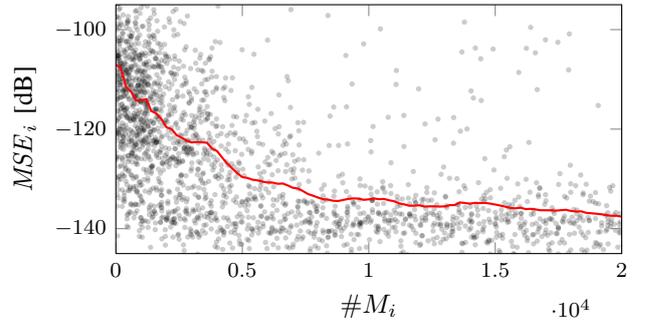}
  \caption{Tracker MSE with respect to number of received measurements.}
  \label{fig:mse}
\end{figure}

The proposed method is applied on each of the $50$ models separately. The
trajectories of the model are sequentially fed into the algorithm. The range of
measurement is assumed to be $[0, 1]$. If the trajectory goes beyond the scope,
it is truncated and no further processing is applied on that. The tracking
performance of the PF is recorded for every trajectory in terms of the MSE
between the ground truth trajectory and the estimation by PF, i.e.
$\mathit{MSE}_i = 10\log \sum (\hat{x}_t^i-x_t^i)^2/T$ for $i^\text{th}$
trajectory. It is expected that over the time the tracker performance improves
as the algorithm updates the learned dynamic model with each measurement. Fig
\ref{fig:mse} shows the scatter plot and the KNN average (red line) of
$\mathit{MSE}_i$ versus the total number of measurements in all the trajectories
received before $i$, i.e. $\#M_i = \sum_{j=1}^{i-1} |\{z^j_0, \cdots \}|$. It is clear
from Fig.  \ref{fig:mse} that by incrementally learning the true dynamic model
the performance of PF significantly improved over $25\text{dB}$.

\begin{figure}[t]
  \setlength\figureheight{3.3cm} 
  \setlength\figurewidth{7cm}
  \input{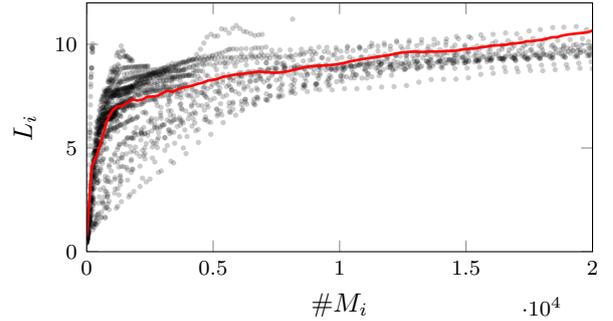}
  \caption{Ground-truth function likelihood with respect to number of received
  measurements.}
  \label{fig:lik}
  \vspace{-10pt}
\end{figure}

Let $L_i = p([f(x_1^*), \cdots f(x_N^*)] |\theta_i)$ be the likelihood of the
ground truth function evaluated on sample point $x_1^*, \cdots, x_N^*$ given the
learned GP model $\theta_i$ up to processing of $i^\text{th}$ trajectory. The
$L_i$ is a relative indication of the closeness of the learned function to the
ground truth function. It is used for evaluating the quality of the incremental
learning algorithm with $N=10^4$ and $x_1^*, \cdots, x_N^*$ uniformly
distributed over $[0, 1]$. Fig \ref{fig:lik} shows the scatter plot of $L_i$
versus $\#M_i$ as well as the KNN average of the values. The empirical
convergence of the proposed method is relatively fast. It averagely converges
with less than $2000$ measurement as shown by Fig. \ref{fig:lik}.

\section {Conclusion}
\label{sec:concl}
A sparse Gaussian process based incremental nonparametric system identification
method for nonlinear state space models is proposed in this paper. The method is
able to update an estimate of the with every measurements from the system.  The
grid inducing point positioning of the proposed method is particularly limits
its usage in high dimensions since lots of the inducing points will placed in
the regions the may not visited by any data. Another limitation of the proposed
method is that due to the underlaying assumption that the dynamics can be model
by function. This will fail when the dynamics is multi modal, i.e.  depending on
some latent effects the dynamic model changes. In future these limitations have
to be addressed.

\clearpage

\bibliographystyle {IEEEbib}
\bibliography {IEEEabrv,bib/refs}

\end {document}